\documentclass[11pt,a4paper]{article}
\usepackage{authblk}
\usepackage[hyperref]{emnlp2020}
\usepackage{times}
\usepackage{latexsym}
\usepackage[leftcaption]{sidecap}

\usepackage{microtype}
\usepackage{booktabs}
\usepackage[ruled, lined, linesnumbered, commentsnumbered, longend]{algorithm2e}
\usepackage{amsmath}
\usepackage{graphicx}

\SetAlCapFnt{\small}
\newcommand{\ROUGE}{\textsc{Rouge}}

\aclfinalcopy %

\title{An Unsupervised Masking Objective for \\ Abstractive Multi-Document News Summarization}

\author{\textbf{Nikolai Vogler}} %
\author{\textbf{Songlin Li}}
\author{\textbf{Yujie Xu}}
\author{\textbf{Yujian Mi}}
\author{\textbf{Taylor Berg-Kirkpatrick}}

\affil{Computer Science and Engineering\\University of California, San Diego \authorcr
  {\tt nvogler@ucsd.edu} }

\date{}

\begin{document}
\maketitle
\begin{abstract}
We show that a simple unsupervised masking objective can approach near supervised performance on abstractive multi-document news summarization.
Our method trains a state-of-the-art neural summarization model to predict the masked out source document with highest lexical centrality relative to the multi-document group.
In experiments on the Multi-News dataset \citep{fabbri-etal-2019-multi}, our masked training objective yields a system that outperforms past unsupervised methods and, in human evaluation, surpasses the best supervised method without requiring access to any ground-truth summaries. Further, we evaluate how different measures of lexical centrality, inspired by past work on extractive summarization, affect final performance.
\end{abstract}

\section{Introduction}
\label{sec:intro}

Multi-document summarization (MDS) aims to condense and combine information from topically related groups of documents into a single concise yet comprehensive short text.
The news domain, where input documents are article clusters of varying size covering the same event, is a common challenge setting for MDS systems \citep{paul2004introduction,owczarzak2011overview}.
While most early approaches to the problem focused on extractive methods that select and copy relevant sections from the input documents \citep{radev2004centroid,erkan2004lexrank,mihalcea2004textrank}, advances in neural text generation have increased both performance of and, subsequently, interest in abstractive methods, which generate a summary from scratch conditioned on a combined representation of the encoded source documents.

\begin{SCfigure*}[1.5][t]
    \centering
    \includegraphics[width=0.7\textwidth]{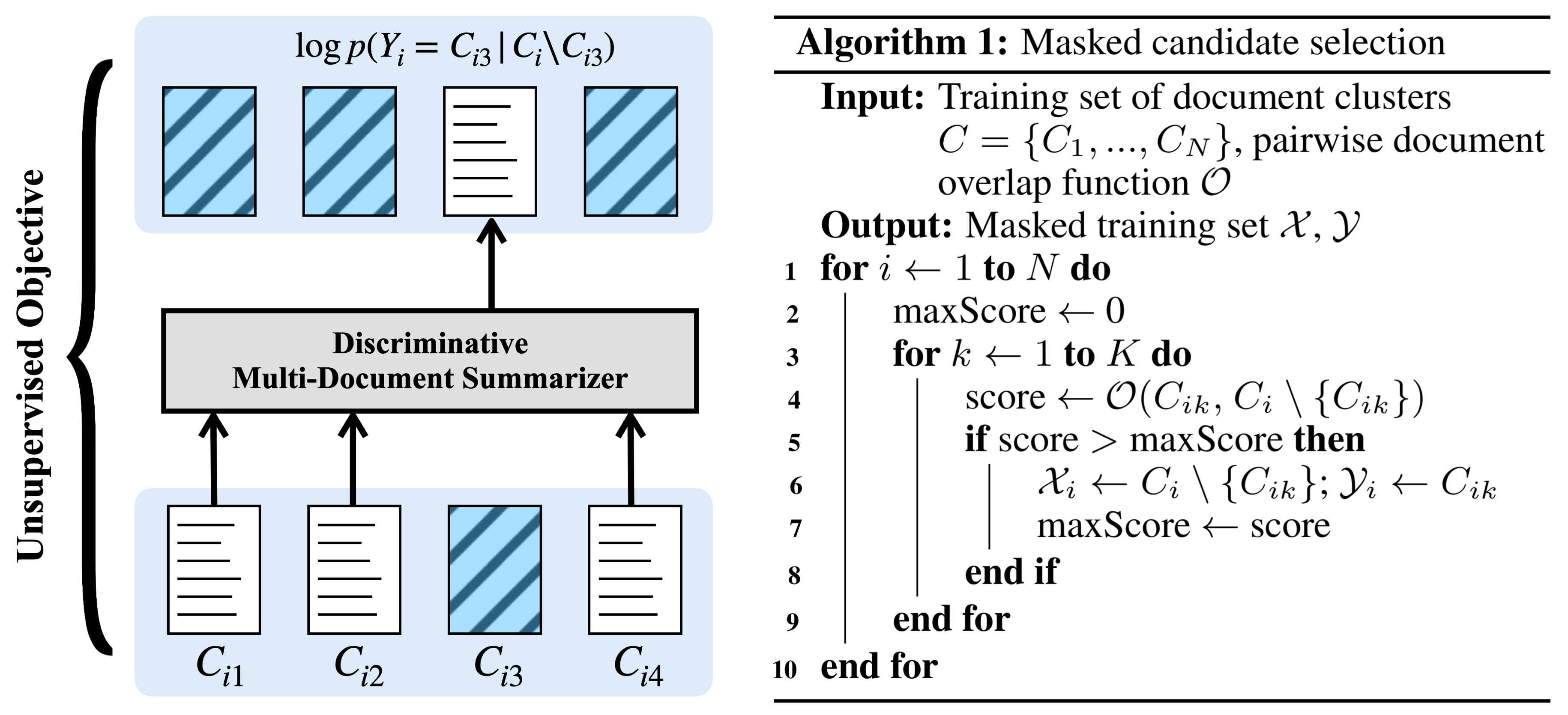}
    \caption{\small Proposed unsupervised masking objective for discriminative MDS models. An input document is selected based on best lexical overlap with the rest of the input documents (in this example, $C_{i3}$). Then a discrimintive summarizer is trained to predict the selected (masked) document conditioned on the rest of the input collection.}
    \label{fig:masking}
\end{SCfigure*}

While large neural systems have seen success on single-document summarization, where human summaries are relatively easy to obtain at scale \citep{napoles-etal-2012-annotated,hermann2015cnndm}, their application to the multi-document task has been restricted by the difficulty of collecting such data---reading and then summarizing a \emph{collection} of documents is a daunting annotation task.
We focus on the task of \emph{unsupervised} abstractive MDS where the system is only allowed access to input document collections at training time \citep{chu2019meansum}.
Similar to older work on unsupervised extractive summarization  \citep{radev2004centroid,erkan2004lexrank,mihalcea2004textrank}, we propose an approach that leverages lexical centrality of the document clusters to select a document to mask out and then predict.
This allows for any, black box discriminative neural abstractive summarization model to be trained without ground-truth summaries by maximizing the likelihood of the masked out document, drawing inspiration from masked language modeling (e.g., BERT \citep{devlin2018bert}).
Some recent works have automatically built large-scale, supervised MDS datasets for news \citep{fabbri-etal-2019-multi} and Wikipedia \citep{liu2018generating}.
We evaluate our unsupervised method on the news domain using the Multi-News dataset \citep{fabbri-etal-2019-multi}.

\section{Masked Document Objective}

Typically, neural abstractive single-document and multi-document summarization models consist of an encoder-decoder architecture \citep{sutskever2014sequence} discriminatively trained end-to-end to maximize the likelihood of the supervised, ground truth summaries given the input document(s).
This can be achieved by minimizing the cross-entropy under the model between the predicted summary's word sequence and the true summary's word sequence.
In order to adapt discriminative training to an unsupervised multi-document setting, we propose a surrogate objective that maximizes the likelihood of a masked document from each topically-related cluster of documents as shown in Figure~\ref{fig:masking}.
Crucially, this procedure can be applied to any black box neural summarization model including current state-of-the-art approaches normally trained on supervised data.

\subsection{Selecting the Best Masking Candidate}

In principle, by training a summarizer to predict randomly chosen masked documents in each input collection, we implicitly train the model to predict the centroid of the collection---a feature of successful extractive systems in past work \citep{radev2004centroid,erkan2004lexrank,mihalcea2004textrank}.
However, since document collections may themselves be noisy, we hypothesize that the summarization model will receive an improved training signal from documents that are more representative of the cluster topic.
By masking the central-most document instead of a random document, we avoid learning to predict diverging documents comprised of irrelevant information.
Contrary to auto-encoding based objectives that learn to reconstruct each input document \citep{chu2019meansum}, our objective encourages overlapping information to be used to predict a target document it does not see as input.
We discuss overlap metrics in the next subsection and outline the candidate mask selection procedure in Algorithm~1.

\subsection{Measuring Document Overlap} %

Ultimately, since {\ROUGE}, a measure based on $n$-gram recall, is summarization's quantitative evaluation measure, we choose to use $n$-gram recall as our overlap metric too. 
Due to both its simplicity and effectiveness, $n$-gram overlap measures have historically been a mainstay for  centroid-based extractive approaches \citep{radev2004centroid,erkan2004lexrank,mihalcea2004textrank}, which measure the lexical overlap between documents of the same topic. 
We empirically investigate using either the unigram or bigram recall between each document in a cluster and the rest of the combined documents in the cluster for the document distance metric, referred to as $\mathcal{O}$ in Algorithm~1.
The document with the highest score is selected to be masked out and predicted.

\section{Related Work}  %
\label{sec:related}

Non-neural unsupervised MDS has been approached from both extractive \citep{carbonell1998use,gillick2009scalable,haghighi2009exploring} and abstractive \citep{mckeown1995generating,radev1998generating,barzilay1999information,ganesan2010opinosis} angles.
At a high level, extractive methods compute a heuristic ranking of source segments and compose them together.
Centroid-based extractive summarization ranks a graph representation of the input document(s), where edges are drawn between segments with similar meanings \citep{radev2004centroid,erkan2004lexrank,mihalcea2004textrank}.
Instead of computing centroid-based score functions to produce an output, in our approach we will use them to select a masked document candidate as an unsupervised objective for training a neural abstractive summarization system.

Supervised neural abstractive single-document summarization has received increased interest lately \citep{rush-etal-2015-neural,see-etal-2017-get,gehrmann-etal-2018-bottom} despite its many  shortcomings \citep{kryscinski2019neural}, such as a reliance on large human-annotated datasets \citep{napoles-etal-2012-annotated, hermann2015cnndm}.
Additionally, supervised abstractive MDS models have also been proposed alongside new datasets 
\citep{liu-lapata-2019-hierarchical,liu2018generating,fabbri-etal-2019-multi}.
Our unsupervised masking objective enables us to train any of the above neural methods discriminatively without ground truth summaries.

Research into unsupervised neural abstractive multi-document summarization is scarcer.
\citet{lebanoff-etal-2018-adapting} and \citet{zhang2018adapting} transferred single-document summarization models to MDS, which required no supervised multi-document data.
\citet{chu2019meansum}'s recently proposed MeanSum, which learns to summarize groups of reviews from a mean-pooled document representation via an auto-encoder objective, represents the most related approach to our own method.

\section{Experiments}
\label{sec:experiments}

\subsection{Dataset}
We use the Multi-News dataset for training and evaluation since it contains supervised summaries for over 125k total news articles in 50k multi-document clusters.
Following \citet{fabbri-etal-2019-multi}, we concatenate each document cluster into a single 500 token mega-document separated by special document separator tokens using the same strategy: for each cluster, the first $500 / M$ tokens are added from each of the $M$ documents, with additional tokens being added iteratively thereafter.\footnote{In practice, we also truncate masked documents to 300 tokens for efficiency.}

\vspace{-.5em}
\paragraph{Cleaning}
Most automatically collected summarization datasets are noisy \citep{kryscinski2019neural} and Multi-News is no exception.
We remove articles less than 100 characters, nonsensical articles containing links to tweets, and duplicated/syndicated articles by thresholding 4-gram overlap. 
Next, we find and delete repeated phrases like captions by computing $n$-gram similarity between sentences within a document.
Finally, we use regular expressions to remove article metadata.
This reduces the dataset from 45K / 5.6K / 5.6K train / valid / test clusters to 42K / 5.2K / 5.2K.

\subsection{Unsupervised Baselines}

We compare to multiple popular unsupervised extractive approaches\footnote{Methods are configured to output similar length summaries to our system.} like \citet{fabbri-etal-2019-multi} along with a recent neural unsupervised approach.

\vspace{-.5em}
\paragraph{Lede-$k$}
We concatenate the first $k$ sentences from each article cluster together to form the summary, which has proven to be a straightforward, but effective baseline for summarization systems \citep{see-etal-2017-get}.

\vspace{-.5em}
\paragraph{LexRank}
\citet{erkan2004lexrank}'s LexRank computes sentence importance via eigenvector centrality in a graph representation of the sentences in each multi-document cluster. 

\vspace{-.5em}
\paragraph{TextRank}
TextRank \citep{mihalcea2004textrank} is another graph-based method that ranks sentences for extraction, originally proposed for single-document summarization.

\vspace{-.5em}
\paragraph{MMR}
Maximal Marginal Relevance \citep{carbonell1998use} ranks candidate sentences based on a balance between relevance and redundancy.
Top-ranked sentences are appended to form an extractive summary.

\vspace{-.5em}
\paragraph{MeanSum}
MeanSum\footnote{\url{https://github.com/sosuperic/MeanSum}} is a completely unsupervised method introduced by \citet{chu2019meansum} for summarizing Yelp and Amazon reviews.
It is trained to simultaneously auto-encode reviews and produce a summary that is semantically similar to the input reviews using their mean-pooled representation.
We remove their review rating classifier and adapt it to the Multi-News dataset.

\paragraph{Source Doc with Best Overlap} We simply return the source document with the highest bigram overlap (i.e., the document that would be masked out and predicted).
This tests how effective our model is at fusing the information between source documents as opposed to just generating a new, separate document on the same topic.

\begin{SCtable*}[1.5][t]
    \small
    \centering
    {\scriptsize
    \begin{tabular}{l | r r | r r | r r | r }
    \toprule
      \ \ \ \ \ \ \ \ \ \ \ \ \ \ \ \ \ \ \ \ \ \ Output Length:& \multicolumn{2}{|c|}{100 Words} & \multicolumn{2}{c|}{150 Words} & \multicolumn{2}{c|}{200 Words} &  \\
      {\bf Method}\ \ \ \ \ \ \ \ \ \ \ \ \ \ \ \ \ \ \ \ \ \ Metric:& R-1 & R-2 & R-1 & R-2 & R-1 & R-2 &  $\mu$ len $\pm$ $\sigma$ \\
\midrule
      \multicolumn{5}{l}{\bf Unsupervised Baselines} \\
    \midrule
      Bigram Overlap Src Doc & 37.35 & 11.39 & 41.02 & 13.08 & 43.29 & 14.25                     & 294 $\pm$ 26 \\
      Random Src Doc &  36.74 & 11.33 & 39.68 & 12.70 & 40.81 & 13.32                             & 203 $\pm$ 71 \\
      Lede-3   &  36.84 & 11.15 & 39.73 & 12.14 & 40.83 & 12.54                                   & 212 $\pm$ 84 \\
      Lede-4   &  36.95 & 11.28 & 40.32 & 12.55 & 42.11 & 13.23                                   & 269 $\pm$ 94 \\
      Lede-5   &  36.98 & 11.34 & 40.47 & 12.74 & 42.56 & 13.56                                   & 318 $\pm$ 98 \\
      LexRank  &  35.30 & 10.28 & 38.85 & 11.93 & 40.89 & 12.85                                   & 246 $\pm$ 15 \\
      TextRank &  36.29 & 11.10 & 39.49 & 12.44 & 41.33 & 13.23                                   & 244 $\pm$ 16\\
      MMR      &  35.62 & 9.35 & 39.89 & 11.48 & 42.57 & 12.91                                   & 237 $\pm$ 17 \\
      MeanSum  &  24.88 & 3.03 & 27.16 & 3.66 & 28.48 & 4.06                                   & 288 $\pm$ 129\\
\midrule
    \multicolumn{5}{l}{\bf Unsupervised Mask Objective} \\  %
\midrule
      Hi-MAP \emph{Random}   &  36.12 & 10.83 & 40.13 & 12.90 & 42.52 & 14.09  &  233 $\pm$ 53\\
      Hi-MAP \emph{Unigram}  &  37.25 & 11.77 & 41.05 & 13.61 & 43.19 & 14.61  &  243 $\pm$  56\\
      Hi-MAP \emph{Bigram}   & 37.21 & 11.48 & 41.01 & 13.33 & 43.35 & 14.54   &  247 $\pm$ 53\\
      Hi-MAP \emph{Random} $\ge 3$   &  36.68 & 11.15 & 40.58 & 13.11 & 42.47 & 13.99  &  213 $\pm$ 46\\
      Hi-MAP \emph{Unigram} $\ge 3$  &  37.03 & 11.28 & 41.00 & 13.31 & 43.28 & 14.47 &  256 $\pm$ 58 \\
      Hi-MAP \emph{Bigram} $\ge 3$   &  \bf 37.58 & \bf 11.60 & \bf 41.45 & \bf 13.63 & \bf 43.75 & \bf 14.82 & 247 $\pm$ 52\\
    \midrule
      \multicolumn{5}{l}{\bf Supervised} & \\
    \midrule
      Hi-MAP & 38.47 & 12.33 & 42.15 & 13.98 & 44.47 & 15.09  & 213 $\pm$ 26\\
      Hi-MAP $\ge 3$ & 37.51 & 11.88 & 41.69 & 14.03 & 43.86 & 15.12  & 199 $\pm$ 35\\
    \bottomrule
    \end{tabular}
    }
    \caption{\small Limited-length {\ROUGE} scores at different output lengths on the Multi-News dataset \citep{fabbri-etal-2019-multi}. The outputs from each model were truncated to meet the length requirement. The distribution of output lengths before truncation for each model is shown in the right-hand column. Results are divided into unsupervised baselines, supervised baselines, and our proposed unsupevised masked objective. The best unsupervised result in each column is \textbf{bolded}. Models with `$\ge3$' in their name indicate that they were only trained on document collections that contained at least three input documents. `Bigram Overlap Src Doc' is a baseline that simply uses the truncated masking candidate as the output document. `Random Src Doc' selects a random input document.}
    \label{tab:results}
\end{SCtable*}

\subsection{Neural Abstractive Summ. Model}
While any neural multi-document abstractive summarization model could be used (e.g., \citet{liu-lapata-2019-hierarchical,liu2018generating}), we use the Hierarchical MMR-Attention Pointer-generator network (Hi-MAP) model \citep{fabbri-etal-2019-multi} because it was developed on the Multi-News dataset.\footnote{\url{https://github.com/Alex-Fabbri/Multi-News}}
As the name implies, the Hi-MAP model is composed of a hierarchical pointer-generator network \citep{see-etal-2017-get} that has an additional attention mechanism over MMR-ranked \citep{carbonell1998use} input sentences.
Thus, sentence-level extractive summarization scores balancing relevancy and redundancy can be effectively combined with a commonly used single document neural summarization model capable of learning when to produce new output or copy from the source vocabulary.
We refer the reader to \citet{fabbri-etal-2019-multi} for hyperparameter settings and implementation details.

\section{Results}
\label{sec:results}

\paragraph{Automatic Evaluation} 
In Table~\ref{tab:results}, we report \ROUGE-\{1, 2\} \citep{lin2004rouge} F-scores for hypotheses\footnote{We use beam search with beam size 4, block repeated trigrams, and apply a length penalty with $\alpha=0.9$ \& $\beta=5$ \citep{wu2016google}.} limited to 100, 150, and 200 words to better compare among uncontrollable neural output lengths (pointed out in \citep{sun2019compare}), along with their untruncated average lengths on the test set summaries from the Multi-News dataset.
Unsupervised methods are purely unsupervised since we use only the masked source documents for the validation set for early stopping.
Since clusters range in size from 2--10 documents in Multi-News, we also experiment with filtering our training set by number of documents per cluster to see if it affects the choice of masks for each unsupervised method.
We report on both the full training set and the other best training set which excludes all 2 document clusters (denoted $\ge3$ in the table).
While {\ROUGE}-2 performance on $\ge3$ is slightly better for the supervised system, filtering out 2-article clusters gives statistically significant gains between models trained with the unsupervised objective.
Allowing the unsupervised objective to choose the article to mask and predict among $\ge3$ articles is clearly important, which is also suggested by the increase in the difference between scores when excluding 2-article clusters from training with a randomly selected masked document and the bigram overlap mask.
We perform a paired bootstrap test on {\ROUGE}-\{1,2\} 200 scores and find that our best method is significantly better than the baselines ($p\ll0.05$) and not significantly different from the best supervised model on {\ROUGE}-1 ($p=0.1$).
We attribute MeanSum's poor performance to it losing supervision from its review rating classifier and the subtle task differences between its original domain (Yelp reviews) and the news domain.

\begin{table}[t]
    \small
    \centering
    \begin{tabular}{l r r r}
    \toprule
        \bf Evaluation Setting            & Info. &     Fluency     & Non-Red. \\
    \midrule
        Ours vs. MeanSum              &    99\% &   99\%  &   95\% \\
        Ours vs. Sup$\ge3$            &    59\% &   53\%  &   50\% \\
    \bottomrule
    \end{tabular}
    \vspace{-0.1in}
    \caption{\small Combined human evaluator preferences for our best system (Bigram$\ge3$) compared to MeanSum and Supervised.}
    \label{tab:human_eval}
    \vspace{-0.2in}
\end{table}

\paragraph{Human Evaluation}
We ask two native English speaking human annotators to evaluate 50 randomly sampled test document clusters and their summaries shuffled between our best bigram method and MeanSum, and our best method and the best supervised Hi-MAP.
Annotators are shown the input documents and either pair of summaries shown above and asked to choose which is \textbf{best} and which is \textbf{worst} \citep{louviere1991best} for informativeness, fluency, and non-redundancy, in line with past work \citep{fabbri-etal-2019-multi}.
Annotators almost always selected our method over MeanSum.
Surprisingly, annotators also ranked our method better than the supervised Hi-MAP system in informativeness (58\%) and fluency (53\%).

\vspace{-0.1in}
\section{Conclusion}
\vspace{-0.1in}
The unsupervised masking method achieves comparable performance with supervised abstractive summarizers. Further, the masking candidate itself proves to be a simple and performant baseline. Similar masking objectives for long-form text generation might find applications in related domains.

\bibliography{emnlp2020}

\begin{thebibliography}{30}
\expandafter\ifx\csname natexlab\endcsname\relax\def\natexlab#1{#1}\fi

\bibitem[{Barzilay et~al.(1999)Barzilay, McKeown, and
  Elhadad}]{barzilay1999information}
Regina Barzilay, Kathleen~R McKeown, and Michael Elhadad. 1999.
\newblock Information fusion in the context of multi-document summarization.
\newblock In \emph{Proceedings of the 37th Annual meeting of the Association
  for Computational Linguistics}, pages 550--557.

\bibitem[{Carbonell and Goldstein(1998)}]{carbonell1998use}
Jaime~G Carbonell and Jade Goldstein. 1998.
\newblock The use of mmr, diversity-based reranking for reordering documents
  and producing summaries.
\newblock In \emph{SIGIR}, volume~98, pages 335--336.

\bibitem[{Chu and Liu(2019)}]{chu2019meansum}
Eric Chu and Peter Liu. 2019.
\newblock Meansum: a neural model for unsupervised multi-document abstractive
  summarization.
\newblock In \emph{International Conference on Machine Learning}, pages
  1223--1232.

\bibitem[{Devlin et~al.(2018)Devlin, Chang, Lee, and
  Toutanova}]{devlin2018bert}
Jacob Devlin, Ming-Wei Chang, Kenton Lee, and Kristina Toutanova. 2018.
\newblock Bert: Pre-training of deep bidirectional transformers for language
  understanding.
\newblock \emph{arXiv preprint arXiv:1810.04805}.

\bibitem[{Erkan and Radev(2004)}]{erkan2004lexrank}
G{\"u}nes Erkan and Dragomir~R Radev. 2004.
\newblock Lexrank: Graph-based lexical centrality as salience in text
  summarization.
\newblock \emph{Journal of artificial intelligence research}, 22:457--479.

\bibitem[{Fabbri et~al.(2019)Fabbri, Li, She, Li, and
  Radev}]{fabbri-etal-2019-multi}
Alexander Fabbri, Irene Li, Tianwei She, Suyi Li, and Dragomir Radev. 2019.
\newblock \href {https://doi.org/10.18653/v1/P19-1102} {Multi-news: A
  large-scale multi-document summarization dataset and abstractive hierarchical
  model}.
\newblock In \emph{Proceedings of the 57th Annual Meeting of the Association
  for Computational Linguistics}, pages 1074--1084, Florence, Italy.
  Association for Computational Linguistics.

\bibitem[{Ganesan et~al.(2010)Ganesan, Zhai, and Han}]{ganesan2010opinosis}
Kavita Ganesan, ChengXiang Zhai, and Jiawei Han. 2010.
\newblock Opinosis: A graph based approach to abstractive summarization of
  highly redundant opinions.
\newblock In \emph{Proceedings of the 23rd International Conference on
  Computational Linguistics (Coling 2010)}, pages 340--348.

\bibitem[{Gehrmann et~al.(2018)Gehrmann, Deng, and
  Rush}]{gehrmann-etal-2018-bottom}
Sebastian Gehrmann, Yuntian Deng, and Alexander Rush. 2018.
\newblock \href {https://doi.org/10.18653/v1/D18-1443} {Bottom-up abstractive
  summarization}.
\newblock In \emph{Proceedings of the 2018 Conference on Empirical Methods in
  Natural Language Processing}, pages 4098--4109, Brussels, Belgium.
  Association for Computational Linguistics.

\bibitem[{Gillick and Favre(2009)}]{gillick2009scalable}
Dan Gillick and Benoit Favre. 2009.
\newblock A scalable global model for summarization.
\newblock In \emph{Proceedings of the Workshop on Integer Linear Programming
  for Natural Langauge Processing}, pages 10--18. Association for Computational
  Linguistics.

\bibitem[{Haghighi and Vanderwende(2009)}]{haghighi2009exploring}
Aria Haghighi and Lucy Vanderwende. 2009.
\newblock Exploring content models for multi-document summarization.
\newblock In \emph{Proceedings of Human Language Technologies: The 2009 Annual
  Conference of the North American Chapter of the Association for Computational
  Linguistics}, pages 362--370. Association for Computational Linguistics.

\bibitem[{Hermann et~al.(2015)Hermann, Ko\v{c}isk\'{y}, Grefenstette, Espeholt,
  Kay, Suleyman, and Blunsom}]{hermann2015cnndm}
Karl~Moritz Hermann, Tom\'{a}\v{s} Ko\v{c}isk\'{y}, Edward Grefenstette, Lasse
  Espeholt, Will Kay, Mustafa Suleyman, and Phil Blunsom. 2015.
\newblock \href {http://arxiv.org/abs/1506.03340} {Teaching machines to read
  and comprehend}.
\newblock In \emph{Advances in Neural Information Processing Systems (NIPS)}.

\bibitem[{Kry{\'s}ci{\'n}ski et~al.(2019)Kry{\'s}ci{\'n}ski, Keskar, McCann,
  Xiong, and Socher}]{kryscinski2019neural}
Wojciech Kry{\'s}ci{\'n}ski, Nitish~Shirish Keskar, Bryan McCann, Caiming
  Xiong, and Richard Socher. 2019.
\newblock Neural text summarization: A critical evaluation.
\newblock \emph{arXiv preprint arXiv:1908.08960}.

\bibitem[{Lebanoff et~al.(2018)Lebanoff, Song, and
  Liu}]{lebanoff-etal-2018-adapting}
Logan Lebanoff, Kaiqiang Song, and Fei Liu. 2018.
\newblock \href {https://doi.org/10.18653/v1/D18-1446} {Adapting the neural
  encoder-decoder framework from single to multi-document summarization}.
\newblock In \emph{Proceedings of the 2018 Conference on Empirical Methods in
  Natural Language Processing}, pages 4131--4141, Brussels, Belgium.
  Association for Computational Linguistics.

\bibitem[{Lin(2004)}]{lin2004rouge}
Chin-Yew Lin. 2004.
\newblock Rouge: A package for automatic evaluation of summaries.
\newblock In \emph{Text summarization branches out}, pages 74--81.

\bibitem[{Liu et~al.(2018)Liu, Saleh, Pot, Goodrich, Sepassi, Kaiser, and
  Shazeer}]{liu2018generating}
Peter~J Liu, Mohammad Saleh, Etienne Pot, Ben Goodrich, Ryan Sepassi, Lukasz
  Kaiser, and Noam Shazeer. 2018.
\newblock Generating wikipedia by summarizing long sequences.
\newblock \emph{arXiv preprint arXiv:1801.10198}.

\bibitem[{Liu and Lapata(2019)}]{liu-lapata-2019-hierarchical}
Yang Liu and Mirella Lapata. 2019.
\newblock \href {https://doi.org/10.18653/v1/P19-1500} {Hierarchical
  transformers for multi-document summarization}.
\newblock In \emph{Proceedings of the 57th Annual Meeting of the Association
  for Computational Linguistics}, pages 5070--5081, Florence, Italy.
  Association for Computational Linguistics.

\bibitem[{Louviere and Woodworth(1991)}]{louviere1991best}
Jordan~J Louviere and George~G Woodworth. 1991.
\newblock Best-worst scaling: A model for the largest difference judgments.
\newblock \emph{University of Alberta: Working Paper}.

\bibitem[{McKeown and Radev(1995)}]{mckeown1995generating}
Kathleen McKeown and Dragomir~R Radev. 1995.
\newblock Generating summaries of multiple news articles.
\newblock \emph{Proceedings, ACM Conference on Research and Development in
  Information Retrieval SIGIR'95}.

\bibitem[{Mihalcea and Tarau(2004)}]{mihalcea2004textrank}
Rada Mihalcea and Paul Tarau. 2004.
\newblock Textrank: Bringing order into text.
\newblock In \emph{Proceedings of the 2004 conference on empirical methods in
  natural language processing}, pages 404--411.

\bibitem[{Napoles et~al.(2012)Napoles, Gormley, and
  Van~Durme}]{napoles-etal-2012-annotated}
Courtney Napoles, Matthew Gormley, and Benjamin Van~Durme. 2012.
\newblock \href {https://www.aclweb.org/anthology/W12-3018} {Annotated
  {G}igaword}.
\newblock In \emph{Proceedings of the Joint Workshop on Automatic Knowledge
  Base Construction and Web-scale Knowledge Extraction ({AKBC}-{WEKEX})}, pages
  95--100, Montr{\'e}al, Canada. Association for Computational Linguistics.

\bibitem[{Owczarzak and Dang(2011)}]{owczarzak2011overview}
Karolina Owczarzak and Hoa~Trang Dang. 2011.
\newblock Overview of the tac 2011 summarization track: Guided task and aesop
  task.
\newblock In \emph{Proceedings of the Text Analysis Conference (TAC 2011),
  Gaithersburg, Maryland, USA, November}.

\bibitem[{Paul and James(2004)}]{paul2004introduction}
Over Paul and Yen James. 2004.
\newblock An introduction to duc-2004.
\newblock In \emph{Proceedings of the 4th Document Understanding Conference
  (DUC 2004)}.

\bibitem[{Radev et~al.(2004)Radev, Jing, Sty{\'s}, and Tam}]{radev2004centroid}
Dragomir~R Radev, Hongyan Jing, Ma{\l}gorzata Sty{\'s}, and Daniel Tam. 2004.
\newblock Centroid-based summarization of multiple documents.
\newblock \emph{Information Processing \& Management}, 40(6):919--938.

\bibitem[{Radev and McKeown(1998)}]{radev1998generating}
Dragomir~R Radev and Kathleen~R McKeown. 1998.
\newblock Generating natural language summaries from multiple on-line sources.
\newblock \emph{Computational Linguistics}, 24(3):470--500.

\bibitem[{Rush et~al.(2015)Rush, Chopra, and Weston}]{rush-etal-2015-neural}
Alexander~M. Rush, Sumit Chopra, and Jason Weston. 2015.
\newblock \href {https://doi.org/10.18653/v1/D15-1044} {A neural attention
  model for abstractive sentence summarization}.
\newblock In \emph{Proceedings of the 2015 Conference on Empirical Methods in
  Natural Language Processing}, pages 379--389, Lisbon, Portugal. Association
  for Computational Linguistics.

\bibitem[{See et~al.(2017)See, Liu, and Manning}]{see-etal-2017-get}
Abigail See, Peter~J. Liu, and Christopher~D. Manning. 2017.
\newblock \href {https://doi.org/10.18653/v1/P17-1099} {Get to the point:
  Summarization with pointer-generator networks}.
\newblock In \emph{Proceedings of the 55th Annual Meeting of the Association
  for Computational Linguistics (Volume 1: Long Papers)}, pages 1073--1083,
  Vancouver, Canada. Association for Computational Linguistics.

\bibitem[{Sun et~al.(2019)Sun, Shapira, Dagan, and Nenkova}]{sun2019compare}
Simeng Sun, Ori Shapira, Ido Dagan, and Ani Nenkova. 2019.
\newblock How to compare summarizers without target length? pitfalls, solutions
  and re-examination of the neural summarization literature.
\newblock In \emph{Proceedings of the Workshop on Methods for Optimizing and
  Evaluating Neural Language Generation}, pages 21--29.

\bibitem[{Sutskever et~al.(2014)Sutskever, Vinyals, and
  Le}]{sutskever2014sequence}
I~Sutskever, O~Vinyals, and QV~Le. 2014.
\newblock Sequence to sequence learning with neural networks.
\newblock \emph{Advances in NIPS}.

\bibitem[{Wu et~al.(2016)Wu, Schuster, Chen, Le, Norouzi, Macherey, Krikun,
  Cao, Gao, Macherey et~al.}]{wu2016google}
Yonghui Wu, Mike Schuster, Zhifeng Chen, Quoc~V Le, Mohammad Norouzi, Wolfgang
  Macherey, Maxim Krikun, Yuan Cao, Qin Gao, Klaus Macherey, et~al. 2016.
\newblock Google's neural machine translation system: Bridging the gap between
  human and machine translation.
\newblock \emph{arXiv preprint arXiv:1609.08144}.

\bibitem[{Zhang et~al.(2018)Zhang, Tan, and Wan}]{zhang2018adapting}
Jianmin Zhang, Jiwei Tan, and Xiaojun Wan. 2018.
\newblock Adapting neural single-document summarization model for abstractive
  multi-document summarization: A pilot study.
\newblock In \emph{Proceedings of the 11th International Conference on Natural
  Language Generation}, pages 381--390.

\end{thebibliography}
\bibliographystyle{acl_natbib}
\end{document}